\documentclass[aip,cha,reprint]{revtex4-1}
\usepackage{graphicx}
\usepackage{color}
\usepackage{epstopdf}
\usepackage{amsmath}
\usepackage{graphicx}
\usepackage{float}
\usepackage{subfigure}
\usepackage{slashbox}
\usepackage[english]{babel}

\begin{document}
\title{Satellite image classification and segmentation using non-additive entropy}

\author{Lucas Assirati} 
\affiliation{Scientific Computing Group, S\~ao Carlos Institute of Physics, University of S\~{a}o Paulo (USP),  cx 369 13560-970 S\~{a}o Carlos, S\~{a}o Paulo, Brazil\\http://www.scg.ifsc.usp.br \\email: assirati@usp.br, bruno@ifsc.usp.br}
\author{Alexandre Souto Martinez} 
\affiliation{Faculdade de Filosofia, Ci\^{e}ncias e Letras de Ribeir\~{a}o Preto (FFCLRP), Universidade de S\~{a}o Paulo (USP), Avenida Bandeirantes, 3900, 14040-901 Ribeir\~{a}o Preto, SP, Brazil - Instituto Nacional de Ci\^{e}ncias e Tecnologia de Sistemas Complexos \\email: asmartinez@ffclrp.usp.br }
\author{Odemir Martinez Bruno}
\affiliation{Scientific Computing Group, S\~ao Carlos Institute of Physics, University of S\~{a}o Paulo (USP),  cx 369 13560-970 S\~{a}o Carlos, S\~{a}o Paulo, Brazil\\http://www.scg.ifsc.usp.br \\email: assirati@usp.br, bruno@ifsc.usp.br}
\begin{abstract}
Here we compare the Boltzmann-Gibbs-Shannon (standard) with the Tsallis entropy on the pattern recognition and segmentation of coloured images obtained by satellites, via ``Google Earth''.
By segmentation we mean split an image to locate regions of interest. 
Here, we discriminate and define an image partition classes according to a training basis.
This training basis consists of three pattern classes: aquatic, urban and vegetation regions. 
Our numerical experiments demonstrate that the Tsallis entropy, used as a feature vector composed of distinct entropic indexes $q$  outperforms the standard entropy. 
There are several applications of our proposed methodology, once satellite images can be used to monitor migration form rural to urban regions, % {\br desmatamento de áreas de preservação ambiental}, 
agricultural activities, oil spreading on the ocean etc.%(???????) 
%The reasons that make the multi index entropy better than the classical one is discussed. 
\end{abstract}


\maketitle

\section{Introduction}

Image pattern recognition is a common issue in medicine, biology, geography etc, in short, in domains that produce huge data in images format. 
Entropy, in its origins is interpreted as a disorder measure. 
Nevertheless, nowadays it is interpreted as the lack of information.
Thus, it has been used as a methodology to measure the information content of a signal or an image. 
In image analysis,  the greater the entropy is, the more irregular and patternless a given image is.  
The additive property of the standard entropy allows its use in several situations just by summing up image characteristics. 
Among the non-additive entropies, we study the Tsallis entropy, which has been proposed to extent the scope of application of classical statistical physics.   
Here, we compare the additive Boltzmann-Gibbs-Shannon (standard)~\cite{Shannon:1948wk} and non-additive Tsallis entropy~\cite{Tsallis:1988ws} when dealing with colored satellite images. 

We start defining the standard entropy for black and white images and we simply extend its use to coloured images, justified by its additive property. 
Next, we consider the Tsallis entropy for black and white images and extend it to coloured images. 
Due to non-additiveness, we call attention to some characteristics that help to qualify these images more efficiently that the standard entropy.   

\section{Non-additive entropy}

Firstly, consider an black and white image with $L_x \times L_y$ pixels. 
The integers $i \in [1,L_x]$ and  $j \in [1,L_y]$ run along the $\hat{x}$ and $\hat{y}$ directions, respectively.  
Let the integer $\tilde{p}_{i,j} \in [0,255]$ represent the image gray levels intensity of pixel $(i,j)$. 
The histograms $\tilde{p}(x)$ of a gray levels image are obtained by counting the number of pixels with a given intensity $\tilde{p}_{i,j}$.

Figure 1 shows an gray scale image and the histogram $\tilde{p}(x)$ produced from this image: 
\begin{figure}[!htbp]
\center
\includegraphics[width=.38\textwidth]{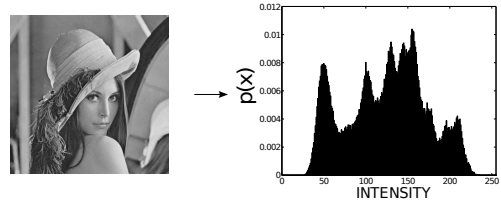}
\label{label}
\caption{Image in gray levels and histogram $\tilde{p}(x)$ }
\end{figure}

To properly use the entropic indexes, one must consider normalized quantities: $p(x) = \tilde{p}(x)/(L_x \times L_y)$, so that normalization condition $\sum_{x=0}^{255}  p(x) = 1$ is satisfied. 
The standard entropy of this image is: \mbox{$H = - \sum_{x=0}^{255}  p(x) \ln p(x) = \sum_{x=0}^{255} p(x) \ln(1/p(x))$}.
\label{eq:black_white_standard_entropy}

Images with low details, produce empty histograms and generate a low entropy value while images with high details produce a better filled histogram, generating high entropy values. Figure 2 illustrates the comparison:

\begin{figure}[H]
\center
\includegraphics[width=.38\textwidth]{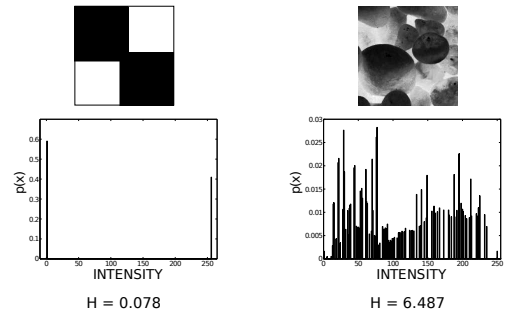}
\label{label}
\caption{Comparison between images with low and high entropy}
\end{figure}

For colored images, a given pixel has three components: red ($k=1$), green ($k = 2$) and blue ($k = 3$), and the integer intensity concerning each one of these colours are written as $\tilde{p}_{i,j,k} \in [0,255]$, so that $k = 1, 2, 3$.
This leads to different histograms for each color: $p_{k}(x)$, and hence different entropies for each color: $H_k$, with $ k = 1,2,3$.  

For two images $A$ and $B$, for a given color, the entropy of the composed image, is the entropy of one image plus the other $H_k(A + B) = H_k(A) + H_k(B)$.
This is the additivity property of the standard entropy, which leads to: 
\begin{equation}
H_k = \sum_{x=0}^{255} p_{k}(x) \ln \left( \frac{1}{p_{k}(x)} \right) \; , k=1,2,3 .
\label{eq:color_standard_entropy}
\end{equation}

Secondly, consider an black and white image mentioned before. 
The Tsallis entropy is for it generalizes the standard entropy\cite{Tsallis:2011hr}: ${S}_{q} = \sum_{x=0}^{255} p(x) \ln_{q}(1/p(x) )$, where the generalized logarithmic function is $\ln_{q}(x) = (x^{q-1} - 1)/(q-1)$,  so that, as $q \rightarrow 1$, one retrieves the standard logarithm, consequently the standard entropy. 

To build a feature vector, one simply uses $n$ different entropic values: \mbox{$\vec{S}_{bw} = (S_{bw,q_1},S_{bw,q_2}, \ldots, S_{bw,q_n}) \; $}, so that $n = 1$ and $q = 1$, one retrieves the standard entropy image qualifier. 
Notice the richness introduced by this qualifier. 
If $n = 1$, we have already an infinity range of entropy indexes to address. 
This richness is amplified for $n > 1$, considering instances of : $q < 1$, $q = 1$ and $q > 1$~\cite{Barbieri:2011gv}. 
 
Since $\ln_{q}(x_1 x_2) = \ln_q(x_1) + \ln_q(x_2) + (1 - q) \ln_{q}(x_1) \ln_q(x_2)$, see Ref.~\cite{Borges:2004gh}, $S_{bw,q}$ is non-additive leading to interesting results when composing two images $A$ and $B$. 
The entropy of the composed image is $S_{bw,q}(A + B) = S_{bw,q}(A) + S_{bw,q}(B) + (1-q) S_{bw,q}(A) S_{bw,q}(B)$, which, for $q \ne 1$ is not simply summation of two entropic values. 
This property leads to different entropic values depending on how one partitions a given image. 
The final image entropy is not simply to summation of the entropy of all its partitions, but it depends on the sizes of these partitions.  

For colored images, we proceed as before, we calculate the entropy of each color component, in principle with different entropy indices values: $q^{(r)}$, $q^{(g)}$ and $q^{(b)}$. 
For sake of simplicity, we consider the same entropic index for all the color components. 
For color $k$ the entropy is:
\begin{equation}
S_q(k) = \sum_{x=0}^{255} p_{k}(x) \ln_q \left( \frac{1}{p_{k}(x)} \right) \; , 
\label{eq:color_tsallis_entropy}
\end{equation} 
so that so that $k = 1, 2, 3$ retrieves Eq.~(\ref{eq:color_standard_entropy}), for $q = 1$. 

\section{Methodology}

Considering pattern recognition in images, the main objective is to classify a given sample according to a set of classes from a database.
In supervised learning, the classes are predetermined. 
These classes can be conceived of as a finite set, previously arrived by a human. 
In practice, a certain segment of data is labelled with these classifications. 
The classifier task is search for patterns and classify a sample as one of the database classes.

To perform this classification, classifiers usually uses a feature vector that comes from a method of data extraction. 
Here, we use the multi-$q$ analyses method \cite{Fabbri2012} \cite{Fabbri2013} that composes a feature vector using certain $q$-entropy values: \mbox{$\vec{S}_q = (S_{q_1}(1); S_{q_1}(2); S_{q_1}(3); \ldots ;S_{q_n}(1); S_{q_n}(2); S_{q_n}(3))$}.

The reason to use the multi-$q$ analysis is that a feature vector gives us more and richer information than a single value of entropy. 
The correct choice of  $q$ indexes emphasize characteristics and provide better classifications.

The following steps describe image treatment, training and validation:
\begin{itemize}
\item Using Google Earth software, capture images from several locations (Figure 3);
\item each image must be segmented in $16 \times 16$ pixels partitions;
\item for each partition the colors Red, Green and Blue are written in a tridimensional array;
\item for each array and for each color, histograms are built and the Tsallis entropies (Eq.~\ref{eq:color_tsallis_entropy}) are calculated, for $q \in [0,2]$ in steps of 1/10;
\item the feature vector is created and the classifiers $k$-nearest neighbors (KNN), Support Vector Machine (SVM) and Best-First Decision Tree (BFTree)  are applied;
\item an output image are delivered with the segmented partitions highlighted (aquatic region = yellow, urban region = cyan, vegetation regions = magenta) according with the classification of KNN classifier.
\end{itemize}

\begin{figure}[!htbp]
\center
\includegraphics[width=.5\textwidth]{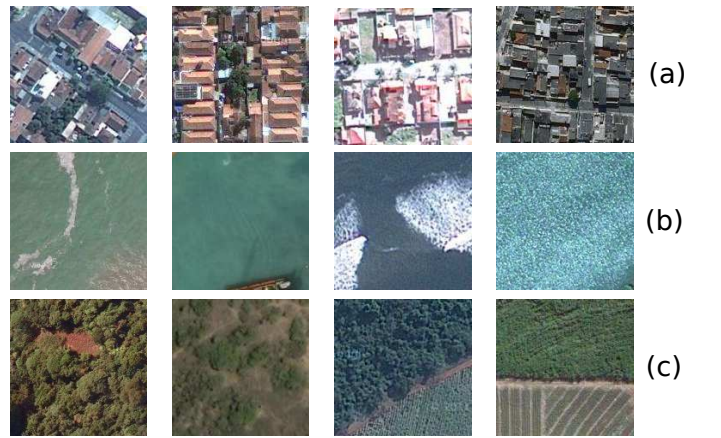}
\label{label}
\caption{Images obtained from Google Earth, from different regions. (a) Urban, (b) Aquatic, (c) Vegetation}
\end{figure}

Figure 4 outlines the steps of the methodology presented:
\begin{figure*}[!htbp]
\center
\includegraphics[width=.8\textwidth]{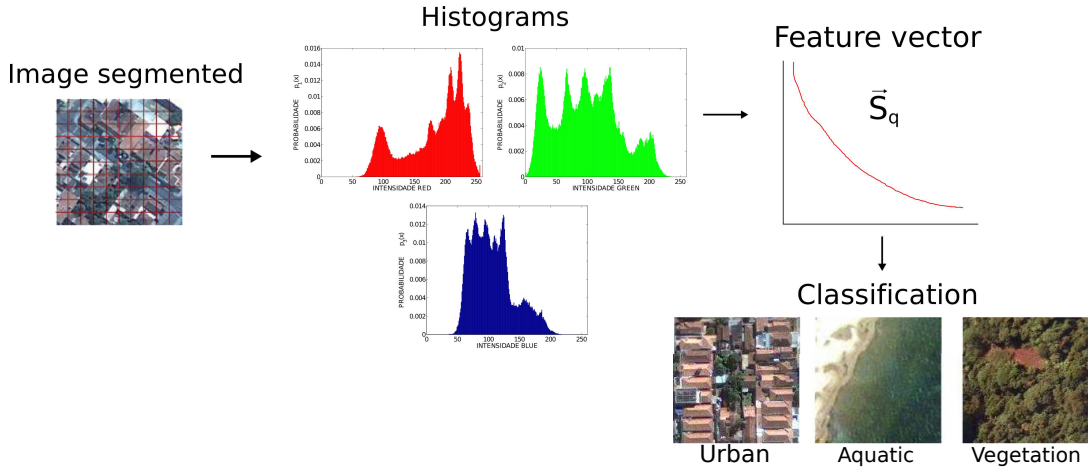}
\label{label}
\caption{schematic drawing of the methodology}
\end{figure*}

Table 1 presents the hit rate percentage of each classifier evaluated for the 3 methods: multi-$q$ analysis, multi-$q$ analysis with attribute selection and standard entropy analysis. 
Since the use of a feature vector gives us more information than a single entropy value it also gives some redundant information. 
In this context, the feature selection is important to eliminates those redundancies.

\begin{table}[!htbp] % aqui comeÃ§a o ambiente tabela
\centering
\caption{Several classifiers are used (SVM, KNN and BFTree) to compare the performance of the generalised entropy with respect to the standard one in pattern recognition. The number of features of each method is indicated in parenthesis.\\{\scriptsize *attribute selection}}% igual ao ambiente figura
\begin{tabular}{|l|r|r|r|} % com este comando dizemos quantas colunas terÃ¡ nossa tabela e a posiÃ§Ã£o do texto dentro de cada coluna. Aqui temos trÃªs colunas (pois sÃ£o trÃªs "c" dentre {}) e o texto estarÃ¡ centralizado em todas elas (indicado pelo "c", se quisermos alinhados Ã  esquerda "l" ou direita "r" 
\hline %Hit rate in percent
\backslashbox{Classifier}{Method} & Multi-$q$ (60) & Multi-$q$ * (8) & BGS (3) \\
\hline % este comando coloca uma linha na tabela
%Classifier         & Multi-q (60) & Multi-q * (8) & BGS (3) \\ % esta Ã© a primeira linha de nossa tabela. O sÃ­mbolo "&" separa as colunas e "\\" indica que aquela linha acabou.
\hline
SVM			&			69.60 $\%$	&		68.96 $\%$	&		65.60 $\%$\\
KNN (1 neighbour)		&			70.80 $\%$	&		69.76 $\%$	&		63.04 $\%$\\
KNN (3 neighbours)		&			72.32 $\%$	&		72.96 $\%$	&		64.00 $\%$\\
KNN (5 neighbours)		&			72.80 $\%$	&		73.28 $\%$	&		64.16 $\%$\\
KNN (7 neighbours)		&			74.88 $\%$	&		72.96 $\%$	&		68.48 $\%$\\
BFTree		&			72.16 $\%$	&		72.80 $\%$	&		67.36 $\%$\\
\hline
\end{tabular}\
\end{table}

Figure 5 depicts image highlights produced by KNN method, evaluated in a region that contains the three types of pattern classes: aquatic, urban and vegetation regions.

\begin{figure}[H]
%\hfill
\subfigure[Original Image]{\includegraphics[width=.22\textwidth]{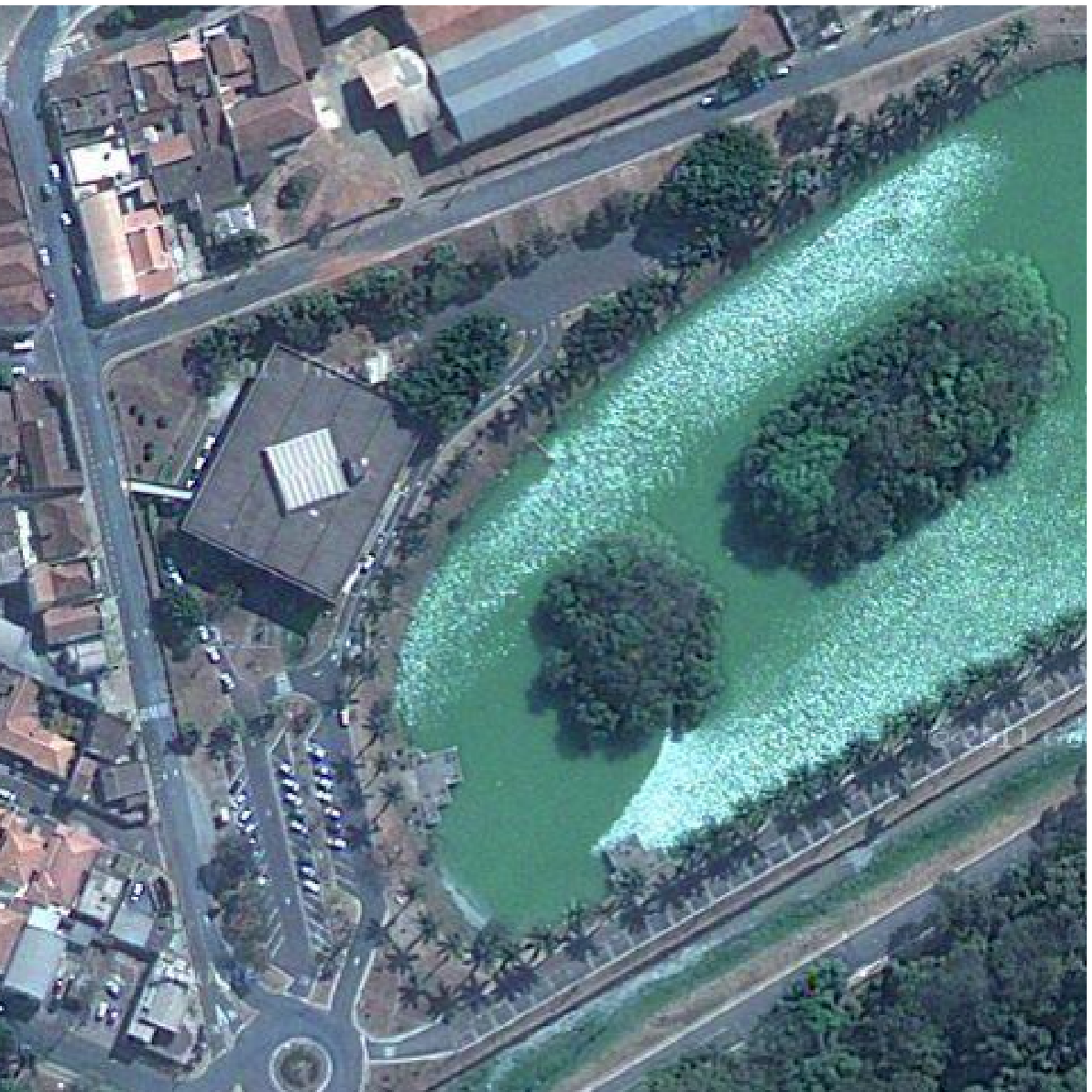}}
%\hfill
\subfigure[KNN with 1 neigbour classifier]{\includegraphics[width=.22\textwidth]{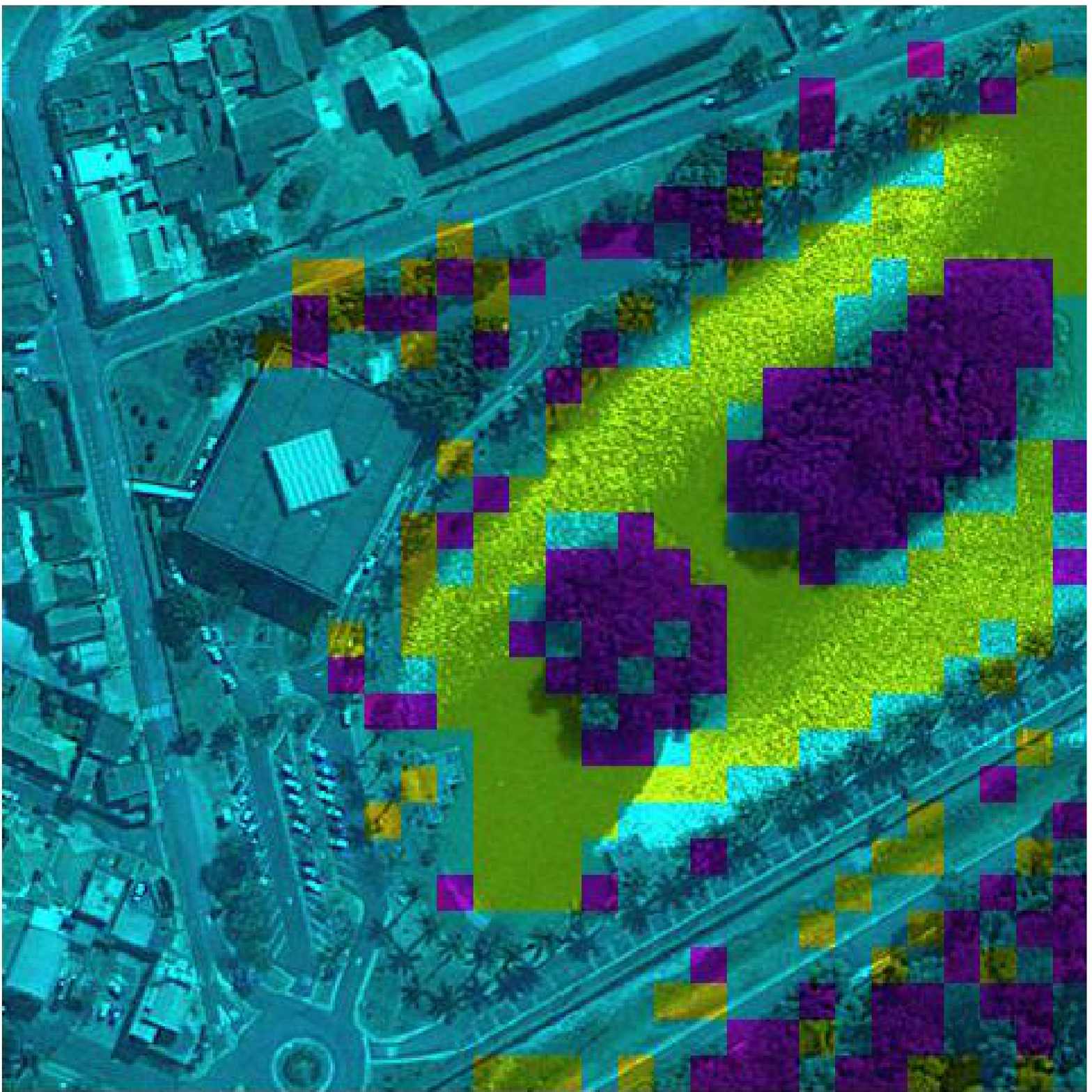}}
%\hfill
\subfigure[KNN with 3 neigbour classifier]{\includegraphics[width=.22\textwidth]{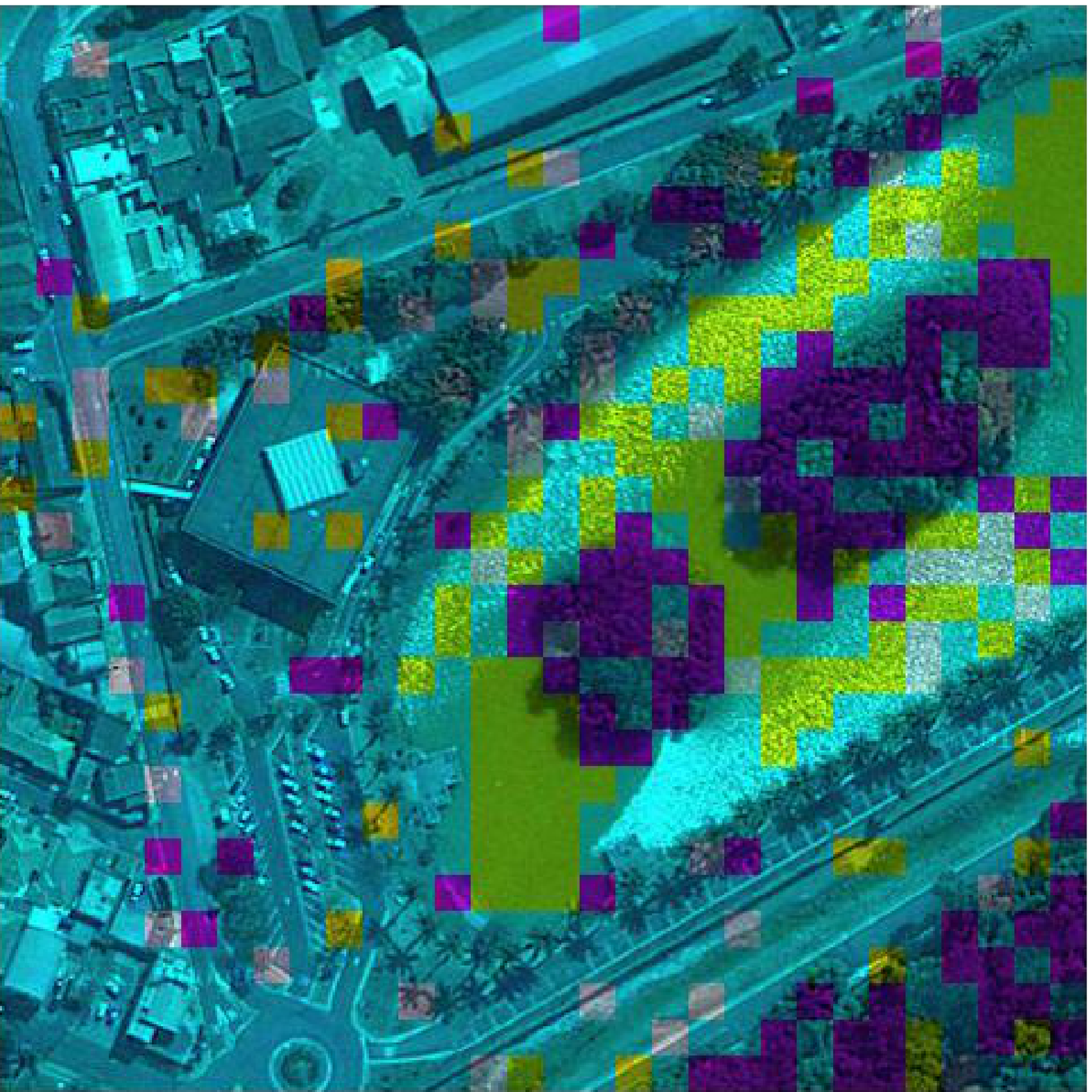}}
%\hfill
\caption{Segmentation obtained by Multi-q method and highlights provided by KNN classifier. The yellow color indicates an aquatic region, the cyan color indicates an urban region and the magenta color indicates a vegetation region. }
\end{figure}

\newpage
\section{Conclusion}
Our study indicates that the Tsallis non-additive entropy can be successfully used in the construction of a feature vector, concerning coloured satellite images.  
This entropy generalizes the Boltzmann-Gibbs one, which can be retrieved with $q = 1$. 
For $q \ne 1$, the image retrieval success is better that the standard case ($q =1$), once the entropic parameter $q$ allows thorougher image exploration. 

\section*{Acknowledgments}
Lucas Assirati acknowledges the Confederation of Associations in the Private Employment Sector (CAPES) Grant . Odemir M. Bruno are grateful for S\~ao Paulo Research Foundation, grant No.:  2011/23112-3. Bruno also acknowledges the National Council for Scientific and Technological Development (CNPq), grant Nos. 308449/2010-0 and 473893/2010-0.

%\section*{References}
%\bibliographystyle{iopart-num} 
%\bibliography{icmsquare}

%merlin.mbs 2010-03-15 4.21a (PWD, AO, DPC)

%\begin{thebibliography}{9}
%\bibitem{iopartnum} IOP Publishing is to grateful Mark A Caprio, Center for Theoretical Physics, Yale University, for permission to include the {\tt iopart-num} \BibTeX package (version 2.0, December 21, 2006) with  this documentation. Updates and new releases of {\tt iopart-num} can be found on \verb"www.ctan.org" (CTAN). 
%\end{thebibliography}

\end{document}